\documentclass[letterpaper, 10 pt, conference]{ieeeconf}
\IEEEoverridecommandlockouts

\usepackage[utf8]{inputenc} 
\usepackage[T1]{fontenc}    
\usepackage{hyperref}       
\usepackage{url}            
\usepackage{booktabs}       
\usepackage{amsfonts}       
\usepackage{nicefrac}       
\usepackage{microtype}      
\usepackage{cite}
\usepackage[font=footnotesize,skip=1pt]{caption}

\usepackage{balance} 
\usepackage{multirow}
\usepackage{color,soul}	
\usepackage{bbm}
\usepackage{bm}
\usepackage{amssymb}
\usepackage{amsmath}
\usepackage{cleveref}
\usepackage{float}

\usepackage{amsmath,amssymb,amsfonts}
\usepackage{algorithm}
\usepackage{algpseudocode}
\usepackage{tikz}
\usepackage{upgreek}
\usepackage{textgreek}
\usepackage{mathtools,mathdots}
\usepackage{xparse}
\usepackage{nameref}
\usepackage{graphicx}
\usepackage{float}
\usepackage{nicefrac}
\usepackage{multirow}
\usepackage{multicol}
\usepackage{hyperref}
\usepackage{multimedia}
\usepackage{arydshln}
\usepackage{substr}
\usepackage{mathrsfs}
\usepackage{pbox}

\let\oldmathcal=\mathcal
\renewcommand{\mathcal}[1]{
    \IfSubStringInString{#1}{ABCDEFGHIJKLMNOPQRSTUVWXYZ}{\oldmathcal{#1}}{
    \IfSubStringInString{#1}{abcdefghijklmnopqrstuvwxyz}{\mathsf{#1}}{
    \ifthenelse{\equal{#1}{\epsilon}}{\textnormal{\straightepsilon}}{
    #1
    }}}
}

\algrenewcommand\alglinenumber[1]{#1:}

\DeclareMathOperator*{\argmin}{argmin}

\newcommand{\reals}{\mathbb{R}}

\newcommand{\Mat}[1][]{\ifthenelse{\equal{#1}{}}{\text{Mat}}{\text{Mat}(#1)}}
\newcommand{\Adjoint}[2]{\textbf{Ad}_{#1}#2}

\newcommand{\SE}[1]{\text{SE}(#1)}

\newcommand{\TSE}[2][]{
	\ifthenelse{\equal{#1}{}}
	{{T\SE{#2}}}
	{{T_{#1}\SE{#2}}}
}
\newcommand{\dualTSE}[2][]{
	\ifthenelse{\equal{#1}{}}
	{{T^*\SE{#2}}}
	{{T^*_{#1}\SE{#2}}}
}

\newcommand{\norm}[1]{\left\lVert#1\right\rVert}

\newcommand{\transpose}{\intercal}

\newcommand{\Vector}[1]{\mathbf{#1}}
\newcommand{\Matrix}[1]{\mathbf{#1}}

\usepackage{soul}
\usepackage{tikz}
\usetikzlibrary{calc}

\makeatletter
\newif\if@anonymize

\@anonymizetrue    

\if@anonymize
  \newcommand{\highlight@DoHighlight}{
    \fill [outer sep = -15pt, inner sep = 0pt, color=black]
          ($(begin highlight)+(0,8pt)$) rectangle ($(end highlight)+(0,-3pt)$) ;
  }

  \newcommand{\highlight@BeginHighlight}{
    \coordinate (begin highlight) at (0,0) ;
  }

  \newcommand{\highlight@EndHighlight}{
    \coordinate (end highlight) at (0,0) ;
  }

  \newdimen\highlight@previous
  \newdimen\highlight@current
  \newlength{\item@width}

  \DeclareRobustCommand*\anonymize{%
    \SOUL@setup
    \def\SOUL@preamble{%
      \begin{tikzpicture}[overlay, remember picture]
        \highlight@BeginHighlight
        \highlight@EndHighlight
      \end{tikzpicture}%
    }%
    \def\SOUL@postamble{%
      \begin{tikzpicture}[overlay, remember picture]
        \highlight@EndHighlight
        \highlight@DoHighlight
      \end{tikzpicture}%
    }%
    \def\SOUL@everyhyphen{%
      \discretionary{%
        \SOUL@setkern\SOUL@hyphkern
        \SOUL@sethyphenchar
        \tikz[overlay, remember picture] \highlight@EndHighlight ;%
      }{%
      }{%
        \SOUL@setkern\SOUL@charkern
      }%
    }%
    \def\SOUL@everyexhyphen##1{%
      \SOUL@setkern\SOUL@hyphkern
      \settowidth{\item@width}{##1}%
      \makebox[\item@width]{}%
      \discretionary{%
        \tikz[overlay, remember picture] \highlight@EndHighlight ;%
      }{%
      }{%
        \SOUL@setkern\SOUL@charkern
      }%
    }%
    \def\SOUL@everysyllable{%
      \begin{tikzpicture}[overlay, remember picture]
        \path let \p0 = (begin highlight), \p1 = (0,0) in \pgfextra
          \global\highlight@previous=\y0
          \global\highlight@current =\y1
        \endpgfextra (0,0) ;
        \ifdim\highlight@current < \highlight@previous
          \highlight@DoHighlight
          \highlight@BeginHighlight
        \fi
      \end{tikzpicture}%
      \settowidth{\item@width}{\the\SOUL@syllable}%
      \makebox[\item@width]{}%
      \tikz[overlay, remember picture] \highlight@EndHighlight ;%
    }%
    \SOUL@
  }
\else
  \newcommand{\anonymize}[1]{#1}
\fi
\makeatother
\newcommand{\radius}{r}
\newcommand{\density}{\rho}
\newcommand{\area}{A}
\newcommand{\length}{L}

\newcommand{\dummyVector}{\Vector{v}}
\newcommand{\position}{x}
\newcommand{\positions}{\Vector{x}}
\newcommand{\director}{\Vector{d}}
\newcommand{\orientation}{\Matrix{Q}}
\newcommand{\rotation}{\Matrix{R}}
\newcommand{\strain}{\epsilon}
\newcommand{\strainCentral}{\epsilon^c}

\newcommand{\curvatures}{\bm{\kappa}}
\newcommand{\curvature}{\kappa}

\newcommand{\shears}{\bm{\nu}}
\newcommand{\shear}{\nu}

\newcommand{\angularVelocities}{\bm{\omega}}
\newcommand{\internalforce}{n}
\newcommand{\internalforces}{\Vector{n}}

\newcommand{\internalcouples}{\Vector{m}}
\newcommand{\externalforces}{\Vector{f}}
\newcommand{\externalcouples}{\Vector{c}}
\newcommand{\transformation}{\gamma}

\newcommand{\springConstant}{k}
\newcommand{\centerToContact}{\bm{\sigma}}
\newcommand{\force}{\Vector{F}}
\newcommand{\couple}{\Vector{C}}

\newcommand{\secondMomentOfInertia}{\Matrix{I}}

\newcommand{\muscle}{\text{m}}

\newcommand{\dataset}{\mathcal{D}}

\newcommand{\lab}[1]{\bar{#1}}

\newcommand{\decisionvariable}{u}
\newcommand{\decisionvariables}{\Vector{\decisionvariable}}

\newcommand{\musclepositions}[1][]{\positions^{\ifthenelse{\equal{#1}{}}{\muscle}{#1}}}
\newcommand{\musclerelativepositions}[1][]{\Vector{\gamma}^{\ifthenelse{\equal{#1}{}}{\muscle}{#1}}}

\newcommand{\cost}{\mathsf{J}}

\newcommand{\musclelength}[1][]{\ell^{\ifthenelse{\equal{#1}{}}{\muscle}{#1}}}
\newcommand{\musclestrain}[1][]{\strain^{\ifthenelse{\equal{#1}{}}{\muscle}{#1}}}
\newcommand{\muscleshears}[1][]{\shears^{\ifthenelse{\equal{#1}{}}{\muscle}{#1}}}
\newcommand{\muscletangent}[1][]{\Vector{t}^{\ifthenelse{\equal{#1}{}}{\muscle}{#1}}}
\newcommand{\muscleforce}[1][]{\internalforce^{\ifthenelse{\equal{#1}{}}{\muscle}{#1}}}
\newcommand{\maxmusclestress}[1][]{\sigma^{\ifthenelse{\equal{#1}{}}{\muscle}{#1}}}
\newcommand{\maxmuscleforce}[1][]{\internalforce^{\ifthenelse{\equal{#1}{}}{\muscle}{#1}}_\text{max}}
\newcommand{\muscleforces}[1][]{\internalforces^{\ifthenelse{\equal{#1}{}}{\muscle}{#1}}}
\newcommand{\musclecouples}[1][]{\internalcouples^{\ifthenelse{\equal{#1}{}}{\muscle}{#1}}}
\newcommand{\muscleactivation}[1][]{a^{\ifthenelse{\equal{#1}{}}{\muscle}{#1}}}
\newcommand{\staticmuscleactivation}[1][]{\alpha^{\ifthenelse{\equal{#1}{}}{\muscle}{#1}}}

\newcommand{\musclestoredenergy}[1][]{W^{\ifthenelse{\equal{#1}{}}{\muscle}{#1}}}

\newcommand{\LM}[1][]{\text{LM}{\ifthenelse{\equal{#1}{}}{}{_{#1}}}}
\newcommand{\OM}[1][]{\text{OM}{\ifthenelse{\equal{#1}{}}{}{_{#1}}}}

\usepackage{tikz}

\newcommand{\BibTeX}{\rm B\kern-.05em{\sc i\kern-.025em b}\kern-.08em\TeX}

\usepackage[normalem]{ulem}

\DeclareRobustCommand{\IEEEauthorrefmark}[1]{\smash{\textsuperscript{\footnotesize #1}}}

\title{Digital twins for the design, interactive control, and deployment \\of modular, fiber-reinforced soft continuum arms.}
\author{
Seung Hyun Kim\IEEEauthorrefmark{1},
Jiamiao Guo\IEEEauthorrefmark{1},
Arman Tekinalp\IEEEauthorrefmark{1},
Heng-Sheng Chang\IEEEauthorrefmark{1,4},
Ugur Akcal\IEEEauthorrefmark{3,4},
Tixian Wang\IEEEauthorrefmark{1,4},\\
Darren Biskup\IEEEauthorrefmark{1},
Benjamin Walt\IEEEauthorrefmark{1},
Girish Chowdhary\IEEEauthorrefmark{3,4},
Girish Krishnan\IEEEauthorrefmark{2},
Prashant G. Mehta\IEEEauthorrefmark{1,4},
Mattia Gazzola\IEEEauthorrefmark{1,5,6,*}
}

\begin{document}
\bstctlcite{IEEEtran} 
\maketitle

\let\thefootnote\relax\footnote{
\IEEEauthorrefmark{1}Mechanical Science and Engineering,
\IEEEauthorrefmark{2}Industrial and Enterprise Systems Engineering,
\IEEEauthorrefmark{3}Computer Science
\IEEEauthorrefmark{4}Coordinated Science Laboratory,
\IEEEauthorrefmark{5}Carl R. Woese Institute for Genomic Biology,
\IEEEauthorrefmark{6}National Center for Supercomputing Applications,
University of Illinois Urbana-Champaign. \IEEEauthorrefmark{*}Corr.: mgazzola@illinois.edu
}

\vspace{-20pt}
\begin{abstract}
Soft continuum arms (SCAs) promise versatile manipulation through mechanical compliance, for assistive devices, agriculture, search applications, or surgery. However, the strong nonlinear coupling between materials, morphology, and actuation renders design and control challenging, hindering real-world deployment. In this context, a modular fabrication strategy paired with reliable, interactive simulations would be highly beneficial, streamlining prototyping and control design. Here, we present a digital twin framework for modular SCAs realized using pneumatic Fiber‑Reinforced Elastomeric Enclosures (FREEs). The approach models assemblies of FREE actuators through networks of Cosserat rods, favoring the accurate simulation of three‑dimensional arm reconfigurations, while explicitly preserving internal modular architecture. This 
enables the quantitative analysis and scalable development of composite soft robot arms, overcoming limitations of current monolithic continuum models. To validate the framework, we introduce a three‑dimensional reconstruction pipeline tailored to soft, slender, small-volume, and highly deformable structures, allowing reliable recovery of arm kinematics and strain distributions. Experimental results across multiple configurations and actuation regimes demonstrate close agreement with simulations. Finally, we embed the digital twins in a virtual environment to allow interactive control design and sim‑to‑real deployment, establishing a foundation for principled co‑design and remote operation of modular soft continuum manipulators.
\end{abstract}

\section{Introduction}

In pursuit of adaptability in complex, unstructured, or delicate environments, soft robotics has remarkably advanced in recent years \cite{Laschi:2016,Yasa:2023}. In particular, soft continuum arms (SCAs), inspired by biological systems such as muscular hydrostats \cite{Kier:1985,Kier:1992} or climbing plants and vines \cite{vaughn20111}, represent compelling solutions towards versatile, comformable, and nimble robotic manipulators \cite{laschi2012soft,Hawkes:2017,Sadeghi:2017,bao2020trunk,Marvi2022Snake,Bas2025MechIntelLocomotion}.
Driven by sustained progress in materials, actuators, and designs \cite{ElAtab:2020,Stano:2021,Sachyani:2021,Zaidi:2021,Marvi2021Material,Bas2024PneumProg,Pikul2022DielectricClutch,Pikul2025VarStiffMat}, SCAs are thus increasingly suited for applications demanding high reconfigurability and safe physical interaction. Examples span minimally invasive surgical procedures \cite{Runciman:2019}, assistive and wearable devices \cite{Bardi:2022,Bas2025AssisDevi}, agricultural manipulation and harvesting \cite{chowdhary2019soft}, or environmental exploration \cite{Milana:2021}.

Despite progress, significant challenges persist \cite{laschi2017soft,Santina2017Challenge,Wang:2018,Su:2022}. A central issue is the inherent tension between increasing mechanical and functional complexity, required to achieve diverse behaviors, and the resulting high-dimensional control space. `Mechanical intelligent' designs \cite{Pfeifer:2006,wang2023mechanical,ulrich1988grasping} can alleviate this tension by exploiting elasticity, structural instabilities, heterogeneity, anisotropy, and environmental interactions to passively handle aspects of arm coordination \cite{hauser2023leveraging,tekinalp2024topology}. However, the systematic development of mechanically intelligent SCAs remains difficult due to the strongly nonlinear coupling between morphology, materials, and control.

Therefore, to support principled design and accelerate innovation, fabrication modularity and predictive, interactive simulation tools are  highly desirable, enabling rapid prototyping, analysis, and structure/control co-design.

\begin{figure}[t!]
\centering
\includegraphics[width=0.47\textwidth]{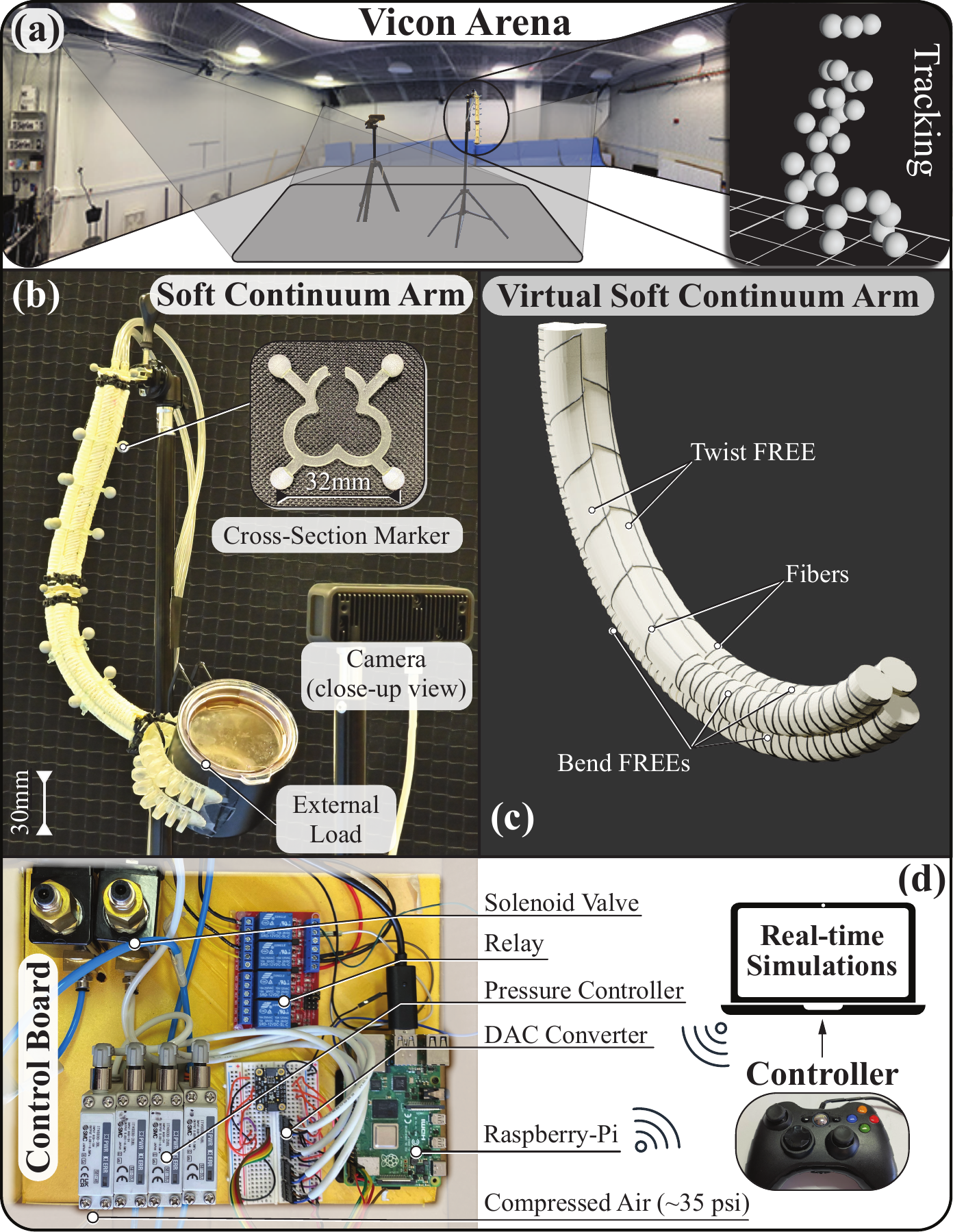}
\caption{\textbf{Setup overview.} 
(a) A soft manipulator \cite{Uppalapati2021BR2} is integrated with the Vicon vision-tracking system \cite{Kim2022Reconstruction}.
(b) The arm is assembled with modular FREE actuators in various serial and parallel configurations, and specifically designed cross-section markers are applied for tracking.
(c) Digital twins assist design and actuation control. Controls are interactively designed using real-time simulations, before deployment to physical hardware.
(d) Computationally identified actuation parameters  are relayed to SMC valves (ITV0031-2UBL) for FREE pressure regulation via Raspberry-Pi and ROS.}
\label{fig:overview}
\vspace{-10pt}
\end{figure}

In this context, we present three main contributions (Fig.~\ref{fig:overview}).

\noindent\textbf{(1)~Digital twins for modular FREE‑based systems.} We developed a quantitative simulation framework that captures the full three‑dimensional dynamics of SCAs modularly realized using pneumatic Fiber‑Reinforced Elastomeric Enclosures (FREEs). The approach is based on \textit{assemblies} of Cosserat rods \cite{Zhang:2019,tekinalp2024topology}, slender elastic elements capable of bending, twisting, stretching, and shearing in 3D space. Each rod is mapped, based on experimental data, to a specialized FREE module and multiple modules are systematically assembled to form complex robotic structures. Although Cosserat rod models have previously been used to describe soft continuum arms \cite{Doroudchi2023SingleCRModel, Wang2024SingleCRModel}, existing approaches typically employ a single, monolithic rod to represent the entire system in a lumped manner, neglecting internal architecture and limiting quantitative accuracy as well as modular and scalable development. Similarly, alternative modeling approaches \cite{Cosimo2020PCC,Armanini:2023,Qin:2024,Chen:2024} have yet to achieve the maturity required to capture the dynamics of assembled soft actuator systems, in which strongly coupled elements interact nonlinearly to generate complex motions. This work is thus the first instance of digital twins for composable FREE-based architectures.

\noindent\textbf{(2)~Experimental verification and integration with the Vicon system.} We developed a three‑dimensional reconstruction pipeline that adapts the Vicon motion‑capture system \cite{Testa2021Vicon,vicon_system} to soft, slender robotic structures. The pipeline overcomes limitations associated with tracking small‑volume, highly deformable bodies, enabling accurate recovery of continuous arm posture and strain distributions. Using this framework, we experimentally validated our simulations across a range of SCA configurations and actuation parameters.

\noindent\textbf{(3)~Interactive control design in virtual environments and sim‑to‑real deployment.} By embedding our digital twins within a virtual environment, we enable the interactive design of manipulation‑oriented control strategies. These controllers are subsequently deployed and demonstrated on physical hardware, providing additional experimental validation and laying the groundwork for remote operation in augmented‑reality, human-in-the-loop settings.

\section{Modular assemblies of FREEs}
\begin{figure*}[t]
\centering
\includegraphics[width=\textwidth]{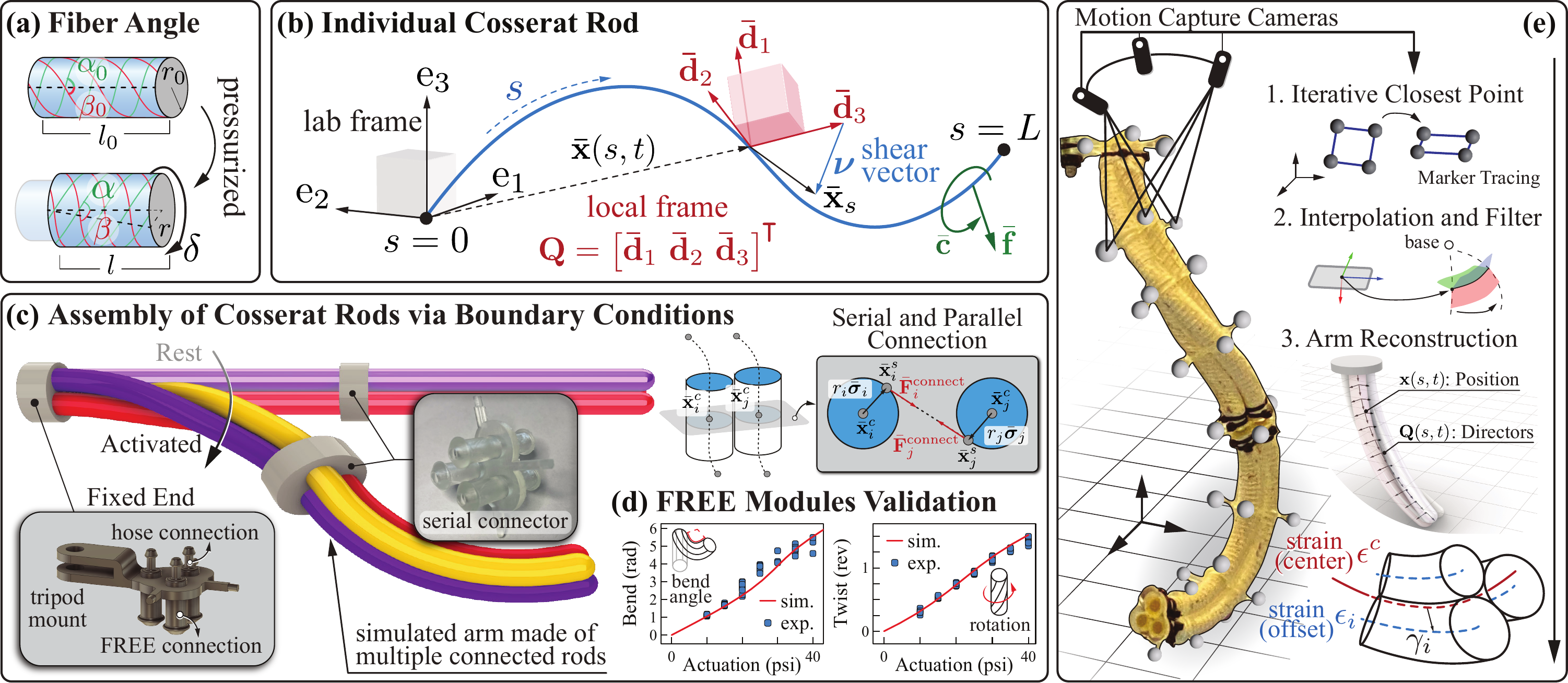}
\caption{\textbf{SCA's modular modeling and reconstruction pipeline.} 
(a) Schematic of FREE fiber angles ($\alpha$, $\beta$) and their change upon pressurization.
(b) Individual Cosserat rod model in 3D space.
(c) Modeling of a BR2-B3 soft arm, composed of six distinct FREEs, using six Cosserat rods assembled via appropriate boundary conditions.
One end of the SCA is fixed (inset shows the base of the arm that connects to pressure regulator).
The two segments (BR2 and B3, each made of three rods/FREEs) are serially connected (inset shows the 3D resin-printed serial connector).
On the right, the schematic illustrates the connection model used in simulations for both parallel and serial connections.
(d) Individual Cosserat rod model replicates primitive deformations of individual FREE modules---bending FREE ($\alpha_{0}=85^\circ$, $\beta_{0}=-85^\circ$) and twisting FREE ($\alpha_{0}=60^\circ$, $\beta_{0}=0^\circ$) of $18~\text{cm}$ lengths.
(e) The reconstruction pipeline for the posture of a SCA in the Vicon system includes the following steps: (1) Iterative Closest Point (ICP) algorithm to track the reflective markers representing cross-sectional planes; (2) SO3-filter \cite{Mahony2005SO3Filter} to denoise both position and directors sequence; (3) reconstruction \cite{Kim2022Reconstruction} to obtain continuous representation of the arm's posture.
}
\vspace{-8pt}
\label{fig:modeling}
\end{figure*}

\textbf{Individual FREE.}
Fiber Reinforced Elastomeric Enclosures (FREE) are pneumatic actuators capable of extending, contracting, bending, and twisting \cite{Krishnan2012FREEdeformation, Krishnan2015FREEdeformation}. 
They consist of a hollow cylindrical enclosure made from a soft elastomer, reinforced with inextensible fibers on the surface. 
When the enclosure expands under pressure, the fibers constrain its deformations, redistributing radial, circumferential, and axial stresses, to give rise to bending, twisting, extension, and contraction in a pre-programmed manner.
Two fibers helically wound at angles $\alpha$ and $\beta$ (Fig.~\ref{fig:modeling}a) lead to twist and either extension (if $\alpha, -\beta > 53.74^{\circ}$) or contraction (if $\alpha, -\beta < 53.74^{\circ}$). 
If $\alpha=-\beta$, the FREE deforms purely axially \cite{Krishnan2015FREEdeformation}.
When an extra $0^\circ$ fiber (elongation limiting spine) is added, bending is attained.

\textbf{Manufacturing FREEs.}  
The fabrication process \cite{Singh2017FREEdeformation,Krishnan2015FREEdeformation} involves wrapping cotton or kevlar fibers on a latex tube using an Arduino-controlled lathe and securing them using rubber cement. 
The winding process may be repeated to increase the helix count, which prevents lateral bulging upon pressurization and favors consistent deformations.
After drying, fresh FREEs are twisted and stretched to eliminate initial hysteresis.
Here, three types of FREEs are considered for modular assembly: extending \& bending ($\alpha = -\beta = 85^{\circ}$ + strain limiting spine), clockwise (CW, $\alpha = 60^{\circ}$, $\beta = 0^{\circ}$) and counterclockwise (CCW, $\alpha = 0^{\circ}$, $\beta = -60^{\circ}$) twisting FREEs.

\textbf{Assembling FREEs.}
Modularity is key to simplify prototyping in both experiments and simulations. FREEs are chosen here because they can be conveniently assembled into complex architectures, to attain elaborate motions via non-linear combinations of deformation modes. We consider coupling FREEs in parallel and serial configurations.

Parallel connectivity is obtained by longitudinally bonding FREEs with rubber cement. If a FREE is a bending actuator, we glue it along its spine. To enhance durability, a latex layer is applied over the bonded area. FREEs are instead serially assembled via a custom, 3D-printed (clear V4 resin from FormLabs, Fig.~\ref{fig:modeling}c) connector fitted to their ends and securely tightened with zip ties (Fig.~\ref{fig:modeling}d). The connector allows the FREEs to share the pressure input, or to retain separate inputs by running a supply line through one of the FREEs.

\textbf{The BR2 soft arm.}
FREE assemblies have been illustrated in worm-like \cite{connolly2015mechanical} or finger-like robots \cite{connolly2017automatic}. 
However, because of the serial arrangement and shared pressure input, motion control is limited and constrained to segmented, planar deformations.
One of the main architectures considered here is instead the BR2 arm, made of three FREEs, one bending and two rotating (twisting) CW and CCW.
Their parallel assembly \cite{uppalapati2018parameter, satheeshbabu2020continuous}, and thus non-linear coupling, conveniently enables complex 3D deformations, increasing workspace while minimizing the number of actuators.

\textbf{The BR2-B3 and B3-BR2 soft arms.}
By further leveraging FREEs modularity, a SCA of expanded workspace is also considered: the BR2-B3 arm \cite{Ripperger2023}.
This consists of two serial segments, each with three FREEs (Fig.~\ref{fig:modeling}c,d). 
The top segment, inspired by the human elbow, features a BR2 manipulator, while the bottom segment, modeled after the human wrist, is composed of three bending FREEs, enabling movement in all directions.
A reversed B3-BR2 configuration is finally used for the interactive control demonstration of Sec.~\ref{sec:interactive-design-and-forward-control}.

\section{Digital twins}

\textbf{Individual Cosserat rod model.} Cosserat rod theory describes the dynamics in 3D space of slender, one-dimensional elastic structures that undergo all modes of deformation (bending, shearing, twisting, stretching) at all cross-sections~\cite{Antman1995Nonelinear, Gazzola:2018}. Individual Cosserat rods naturally map to individual FREEs (which are slender by design) allowing to model FREEs non-linear mechanics, material properties, active stresses, and environmental loads \cite{Krishnan2012FREEdeformation, Krishnan2014FREEdeformation, Krishnan2015FREEdeformation}.

A Cosserat rod (Fig.~\ref{fig:modeling}b) is described by its centerline $\lab{\positions}(s,t) \in \reals^{3}$ and orthonormal reference frame $\orientation(s,t) \in \reals^{3 \times 3} =\left[\lab\director_{1}, \lab\director_{2}, \lab\director_{3}\right]^{\transpose}$, leading to the relation for any vector $\dummyVector: \dummyVector = \orientation \lab\dummyVector$, $\lab\dummyVector = \orientation^{\transpose}\dummyVector$, where $\lab\dummyVector$ denotes the vector in the lab frame and $\dummyVector$ in the local frame. Here, $s \in [0, \length]$ is the rod arc-length and $\length(t)$ its current length, for time $t \in \reals \ge 0$. For inextensible, unshearable rods, $\lab\director_{3}$ is parallel to the tangent ($\partial_s \lab\positions = \lab\positions_{s}$), with  $\lab\director_{1}$, $\lab\director_{2}$ spanning the normal-binormal plane. Shear and extension shift $\lab\director_{3}$ away from $\lab\positions_{s}$, producing the shear strain vector $\shears = \orientation \left(\lab\positions_{s} - \lab\director_{3} \right)$. The curvature vector $\curvatures$ encodes bending/twist and is the $\orientation$'s rotation rate along the material coordinate $\partial_s \director_j = \curvatures \times \director_j$, while the angular velocity $\angularVelocities$ is $\partial_t \director_j = \angularVelocities \times \director_j$. The linear velocity of the centerline is $\partial_t\lab\positions$, the second area moment of inertia is $\secondMomentOfInertia(s,t)\in \reals^{3 \times 3}$, the cross-sectional area is $\area(s,t)\in \reals$, density is $\density(s)\in \reals$. Then, the governing equations are \cite{Gazzola:2018} 
\begin{equation}
    \partial_{t} \left( \density \area \partial_{t} \lab\positions \right)= \partial_s \left(\orientation^{\transpose} \internalforces\right) + \lab\externalforces
    \label{eqn:linear_momentum}
\end{equation}
\begin{equation}
    \partial_t \left( \density \secondMomentOfInertia \angularVelocities \right) = \partial_s \internalcouples + \curvatures \times \internalcouples + \left(\orientation \lab\positions_s \times \internalforces\right)
     + \left(\density \secondMomentOfInertia \angularVelocities\right) \times \angularVelocities + \orientation\lab\externalcouples
     \label{eqn:angular_momentum}
\end{equation}
where $\internalforces(s,t)\in \reals^{3}$ and $\internalcouples(s,t)\in \reals^{3}$ are internal forces and couples, respectively, that depend on elastic deformations, geometry, and material constitutive model. Finally, $\lab\externalforces(s,t)\in \reals^{3}$ and $\lab\externalcouples(s,t)\in \reals^{3}$ are the cumulative external forces and couples (per unit length) applied to the rod. Here, external loads are used to capture rod-to-rod connectivity (see next), gravity, and Hertzian contact \cite{Hertz:1882}, as detailed in \cite{tekinalp2024topology}.

\textbf{Assembling Cosserat rods.} 
To form rod assemblies representative of the complex mechanics of FREE-composed arms, boundary conditions that mimic parallel `gluing' or serial connectivity are necessary. Mathematically, these amount to ensuring that glued surfaces or rigidly connected ends share the same positions in space at all times, while maintaining relative orientations. Here, we employ load-displacement relations to produce restoring forces and couples that preserve physical connectivity \cite{Zhang:2019, tekinalp2024topology}.

Let us consider (Fig.~\ref{fig:modeling}c) two connected elements from two distinct rods. The surface location of element $i$ in contact with
the neighboring element $j$ is defined as $\lab\positions^{s}_{i}=\lab\positions^{c}_{i}+\radius_{i} \lab\centerToContact_{i}$, where $\lab\positions^{s}_{i}$, $\lab\positions^{c}_{i}$, and $\radius_{i}$ are the element's center position, contact position, and radius respectively, and $\lab\centerToContact_{i}$ is the center-to-contact unit vector.
In the rest configuration, $\lab\centerToContact_{i}$ points from $\lab\positions^{c}_{i}$  to $\lab\positions^{c}_{j}$ and $\lab\centerToContact_{j} = -\lab\centerToContact_{i}$.
Then, at each timestep, the restoring force $\lab\force^\text{connect}_{i} = \springConstant_{s} \left(\lab\positions^{s}_{j} - \lab\positions^{s}_{i}\right)$ and couple $\lab\couple^\text{connect}_{i} = r_{i} \lab\centerToContact_{i} \times \lab\force^\text{connect}_{i}$, is applied to each element. Here, the force spring constant $\springConstant_s$ is chosen to prevent separation and avoid numerical instabilities. 

Next, we need to maintain the relative orientation between two connected elements. First, we define $\orientation_{i}$ for element $i$ and $\orientation_{j}$ for element $j$. 
The rotation matrix $\rotation_{ij} := \orientation_{j}\orientation_{i}^{\transpose} \in \reals^{3 \times 3}$, where $\intercal$ denotes the matrix transpose, transforms $\orientation_{i}$ into $\orientation_{j}$.
In the rest configuration (hat symbol), the same rotation matrix reads $\hat{\rotation}_{ij} := \hat{\orientation}_{j} \hat{\orientation}_{i}^{\transpose}$.
The torque-orientation relation considers the deviation between the current orientation $\rotation_{ij}$ and the rest configuration $\hat{\rotation}_{ij}$ to compute the orientation-restoring couple $\lab\couple^\text{align}_i = -\springConstant_{t} \orientation_{j}^{\intercal} \log\left(\hat{\rotation}_{ij}^{\transpose}\rotation_{ij}\right)$, where $\springConstant_t$ is chosen to maintain alignment and numerical stability.
The operator $\log(\cdot) : \reals^{3 \times 3} \to \reals^{3}$ is the matrix logarithm, which converts a rotation matrix into its corresponding rotation vector. 
Equal and opposite torques are applied to neighboring element $j$ $(\lab\couple^\text{align}_{i} = -\lab\couple^\text{align}_{j})$.

\textbf{Digital twins -- Cosserat rods to FREEs.}
To realize digital twins of FREE-based systems, rods' material and geometric properties are quantitatively matched with FREE modules (Table~\ref{table:param}). Further, the effects of pressurization and fiber inextensibility that govern individual FREEs' deformations are quantitatively couched into our rods as resultant loads.

To determine them, we define the FREE's axial $\lambda_{1} := l/l_{0}$ and radial $\lambda_{2} := r/r_{0}$ stretch, where $l$, $l_{0}$ and $r$, $r_{0}$ are the current and initial lengths and radii, respectively. As the FREE deforms under pressure (Fig.~\ref{fig:modeling}a), fibers' inextensibility leads to changes in the winding angles \cite{Krishnan2012FREEdeformation}. The kinematic equations governing these changes are $\alpha = \tan^{-1} \left(\frac{\lambda_{2}}{\lambda_{1}}\tan(\alpha_{0}) + \frac{r}{l}\delta\right)$ and $\beta = \tan^{-1} \left(\frac{\lambda_{2}}{\lambda_{1}}\tan(\beta_{0}) + \frac{r}{l}\delta\right)$, where $\alpha_{0}$ and $\beta_{0}$ are the fiber angles in the rest state, and $\delta$ is the twist of the FREE in its deformed configuration.

To connect these kinematic equations with the rods' dynamics, we employ the relationships described in \cite{Bishop-Moser2013FREEactuation} that estimate the effective internal forces and couples acting on the FREE given fiber angles $\alpha$, $\beta$ and fluid pressure $P$
\begin{align}
    F &= \gamma P \pi r^2 \left(1 + 2 \cot(\alpha)\cot(\beta)\right) \nonumber \\
    &\frac{
    S(\alpha)^2S(\beta)^2S(\alpha-\beta)^2
    }{S(\alpha)^2S(\beta)^2S(\alpha-\beta)^2+\left(S(\alpha)^2 - S(\beta)^2\right)^2}
\end{align}
\vspace{-10pt}
\begin{align}
    C &= \gamma P \pi r^{3} \left(1 + 2 \cot(\alpha)\cot(\beta)\right) \nonumber \\
    &\frac{ 
     S(\alpha)S(\beta) S(\alpha-\beta)\left(S(\alpha)^2-S(\beta)^2\right)
    }{S(\alpha)^2S(\beta)^2S(\alpha-\beta)^2 + \left(S(\alpha)^2 - S(\beta)^2\right)^2}
\end{align}
\noindent
where $\gamma$ is introduced to account for manufacturing variability ($\pm 10\%$), and $S(\cdot)=\sin(\cdot)$. The effective loads so computed are applied to our rods as $\lab\force^\text{load} = F\lab\director_{3}$ along the tangent direction, and $\lab\couple^\text{load} = C\lab\director_{3}$ along the twist direction. These relations do not consider the presence of a spine, and therefore do not model bending FREEs, only extending/shortening and twisting ones. However, we recall here that the spine (additional inextensible $0^\circ$ fiber) constraints extension on one side of the FREE/rod, giving rise to the additional bending couple $\lab\couple^\text{bend} = \mu\left(r\lab\director_{f} \times F\lab\director_{3} \right)$, where $r$ is the rod's radius, $\lab\director_{f}$ is the unit vector (in the normal-binormal plane) pointing from the centerline to the spine, and $\mu$ is a scaling parameter calibrated based on experiments. All model parameters for all simulations are reported in the Appendix.

\textbf{\textit{Elastica} software.} 
With a closed system of equations, we proceed with its numerical integration using \textit{Elastica} \cite{PyElastica}, whose accuracy has been demonstrated via rigorous benchmarks \cite{Gazzola:2018}, quantitative biomechanical simulations \cite{Zhang:2019, Zhang:2021, porat2023mechanical, tekinalp2024topology}, and for the design/control of artificial  \cite{Naughton:2021, tekinalp2024topology, Chang:2020, shih2023hierarchical} and bio-hybrid \cite{Zhang:2019,kim2023remote,Wang:2021} soft robots.

\begin{figure*}[t]
\centering
\includegraphics[width=\textwidth]{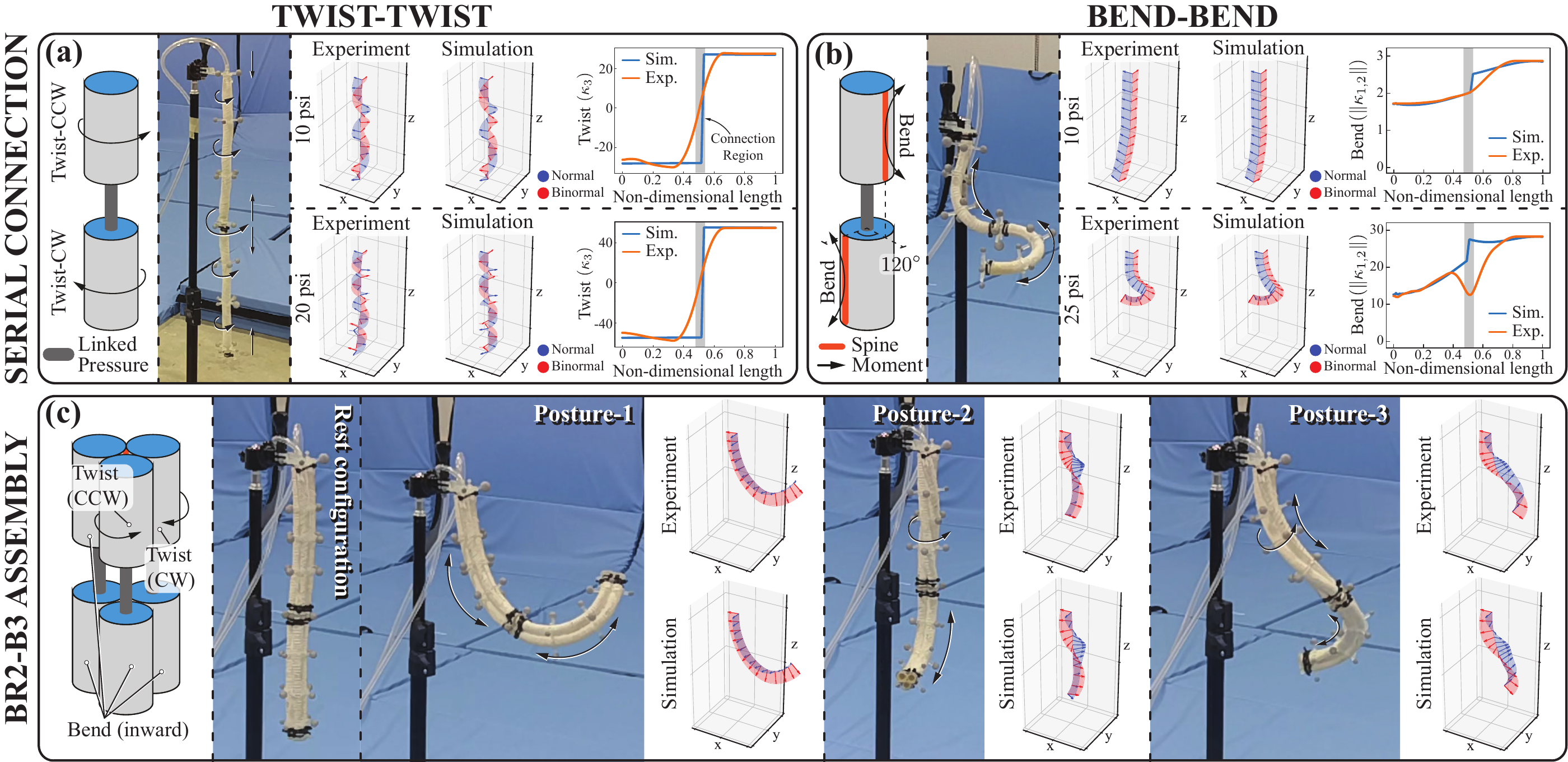}
\caption{
\textbf{Experimental validation.} 
For all simulations, we use the Young’s modulus of the elastomer FREEs are made of, and match the area $(A)$ and moment of inertia $(I)$ of each individual rod with the corresponding hollow FREE. Rod modules are then connected together to form various configurations, which are subsequently fabricated for experimental verification.  
(a) Twist-twist and (b) bend-bend validations are conducted for serially connected FREEs. 
The validation includes a comparison of the centerline position and orientation (normal and binormal directions) between experiments and simulations, together with the curvature profile (twist: $\curvature_3$, bend: $\|\curvature_{1,2}\|=|(\curvature_1,\,\curvature_2)|$) along the arm’s length. The connection region is shaded in grey in the plots.
(c) Three different actuation combinations of the BR2-B3 SCA are tested for validation. These include: (posture-1) actuation of two serially connected bending FREEs at 25 psi, (posture-2) actuation of serially connected bending and twisting FREEs at 20 psi, and (posture-3) a combined actuation of (posture-1) and (posture-2).
For the BR2-B3 validation, we compare the posture between the experimental results and simulations. Errors between simulated and experimental postures and strains are listed in Table I, together with the corresponding used metrics. In all cases, deviations are found to be less than 10\%.
}
\vspace{-8pt}
\label{fig:validation}
\end{figure*}

\section{Experimental validation}

\textbf{The Vicon arena}. The Vicon motion capture system \cite{Testa2021Vicon, vicon_system} is one of the most advanced video tracking and interpretation solutions commercially available.
Widely used \cite{van2019agreement, pfister2014comparative, adesida2019exploring} in applications ranging from virtual reality and entertainment to biomechanics and engineering, it provides precise tracking of 3D body motions at $120~\text{Hz}$ frame rate, by means of specialized multiple cameras and reflective markers placed on the subject’s body.

The Vicon system tracks `objects', defined as collections of markers whose relative positions are assumed to remain fixed over time. By design, robust identification is limited to objects that are rigid, geometrically distinct, and spatially well separated---conditions that do not necessarily well align with our application. Indeed, our soft robotic arms undergo small-volume, spatially localized deformations relative to the Vicon work volume, yet exhibit complex motions that require dense, closely spaced marker-objects (to estimate orientations, see Fig.~\ref{fig:overview}b) along the robot length. During actuation, twisting and bending further bring markers into close proximity. Under these conditions, the system is prone to common failure modes, including marker flickering, swapping, dropout or ghosting as well as incorrect object labeling. These errors lead to reduced measurement accuracy and compromise the reliable reconstruction of arm postures and strain fields.

To mitigate these issues, we replace the built-in object tracking algorithm with three custom pre-processing steps on the Vicon datastream (Fig.~\ref{fig:modeling}e). First, we implement the Iterative Closest Point (ICP) algorithm \cite{Neil1992ICP,Nora2017ICP} and employ it to identify and group the path trajectories for each cross-section marker (Fig.~\ref{fig:overview}b), so as to trace the position and director of every cross-section. Once the trajectory of the position and director of each cross-section are measured, a SO3-filter \cite{Mahony2005SO3Filter} is used to eliminate high-frequency jitter in the orthogonal matrices. Finally, a continuous representation of the arm's posture and strains is reconstructed based on the discrete data, following the physics-informed reconstruction approach detailed next. As demonstrated in the remainder of the paper, our overall reconstruction methodology is found to be effective, rendering the powerful capabilities of the Vicon systems fully available to soft slender body research.

\textbf{Posture and strain reconstruction.} 
To accurately reconstruct posture (centerline position and orientation) and all deformation modes (bending, shearing, twisting, stretching) of our SCAs, we employ a physics-informed method tailored to soft, slender bodies \cite{Kim2022Reconstruction}. This technique does not consider only kinematic data, as most approaches, but also incorporates the mechanics of the soft body to discount non-physical or highly energetic body configurations. This is attained by couching the reconstruction process as an optimization problem where the arm centerline's posture/strains are identified by minimizing both mechanical potential energy and data mismatch relative to experimental measurements. Concisely,
\begin{equation}\label{eq:smoothing_problem}
\min_{\decisionvariables}~\cost(\decisionvariable;\dataset),~\text{s.t. kinematic \& smoothness constraints}
\end{equation}
\noindent where $\decisionvariable$ is the decision variable that controls the smoothness of the strain, $\dataset$ is the marker data, $\cost$ is the objective function that contains both energy and data mismatch costs.
This approach, which has been previously demonstrated for slender soft structures and whose detailed formulation can be found in \cite{Kim2022Reconstruction}, enables non-invasive, high-accuracy, and comprehensive posture/strain estimation in the present context.

As noted, the above method applies to the centerline of the experimental arm. To compare with simulations (where assembled rods may be at an offset distance from the arm's centerline, Fig.~\ref{fig:modeling}c), we need to map the posture and strains of each rod to the centerline of the arm. To this end, we employ an averaging optimization technique \cite{Cheng2007AvgQuat} for which $\strainCentral = \argmin_{\strain}~\norm{\tfrac{1}{N}\sum_{i=1}^N(\strain_i - \Adjoint{\transformation_i}{(\strain}))}$.
Here, $\strain_i$ is the strain of the $i^\text{th}$ simulated rod and $\transformation_i$ is the transformation matrix (Fig.~\ref{fig:modeling}e) that maps the positions and orientations of the $i^\text{th}$ rod to the arm's centerline. The optimization seeks to minimize the distance between the adjoint $\Adjoint{\transformation_i}{(\strain)}$, which represents the strain of the $i^\text{th}$ rod estimated from the unknown centerline strain $\strain$, and the actual simulated strain $\strain_i$. The result of this minimization process is the centerline strains $\strainCentral$, from which the posture is derived, allowing us to directly compare experimental and computational results. 

\textbf{Validation.}
With the machinery in place to perform and compare simulations and experiments, we proceed with its verification. Simulation parameters are first calibrated based on individual FREEs deformation primitives (Fig.~\ref{fig:modeling}d), providing the computational building-blocks for all subsequent tests and benchmarks with composite SCAs.
We then begin with a case involving a SCA composed of two serially connected twisting FREEs, one rotating clockwise (CW) and the other counterclockwise (CCW) under pressure (Fig.~\ref{fig:validation}a).
Due to the opposing twist directions, the twist reverses at the center of the SCA, resulting in the largest rotation angle at the serial connector's location. This reversal poses a challenge to the torque-orientation connection model, as it must precisely match the end orientations at the rods' meeting point. We test two different actuation pressures and compare simulations and experiments against the SCA's posture and twist profile.
As can be seen in Fig.~\ref{fig:validation}a, computations reproduce experimental measurements of twist profile and magnitude, capturing the reversal at the connector location as well as the overall arm posture.
Observed (small) deviations are quantified in Table~\ref{tab:accuracy}, and attributed to manufacturing, material, and experimental variability as well as modeling assumptions in both reconstruction algorithm and simulations. For all cases reported in this work, discrepancies are found within the 2-8\% range, as detailed in the Appendix.

Next, we benchmark two serially connected bending FREEs, presenting a $120^{\circ}$ angle between their spines. This arrangement causes, upon pressurization, the segments to bend in different planes with opposite concavity, giving rise to the 3D morphology of Fig.~\ref{fig:validation}b. This is a challenging test. However, our simulations recapitulate the experimentally reconstructed arm's posture, curvature profile, and magnitude. We emphasize how, at 25~psi, the arm deforms rather dramatically compared to 10~psi, highlighting the ability of our simulations to capture this transition. However, we note the mismatch at the connection region. This is due to a slight misalignment between the spines of the two FREEs, but also because of the jump conditions of the reconstruction algorithm, introduced for the connecting components.

Finally, we consider the BR2-B3 arm across a set of postures corresponding to different pressurization settings (Fig.~\ref{fig:validation}c).
Given the symmetric design and actuation space, bend-bend and clockwise bend-twist actuation combinations are presented.
This serves as a comprehensive test, since the parallel and serial assembly of six distinct FREEs can produce a number of deformation modes.
In Posture 1, we actuate two bending FREEs, one in each serial segment (BR2 and B3). The two active FREEs are assembled with aligned spines to allow the arm to purely bend in-plane. As seen in Fig.~\ref{fig:validation}c, the model and the physical arm recover this pre-programmed behavior.
In Posture 2, we pressurize a twisting FREE in the top segment and a bending FREE in the bottom segment (Fig.~\ref{fig:validation}c).
Lastly, we combine the actuations from Posture 1 (25~psi) and Posture 2 (20~psi), driving the arm into a 3D helical configuration.
Once again, our model is found to capture the observed experimental arm's reconfigurations.

\begin{figure*}[t]
\centering
\includegraphics[width=\textwidth]{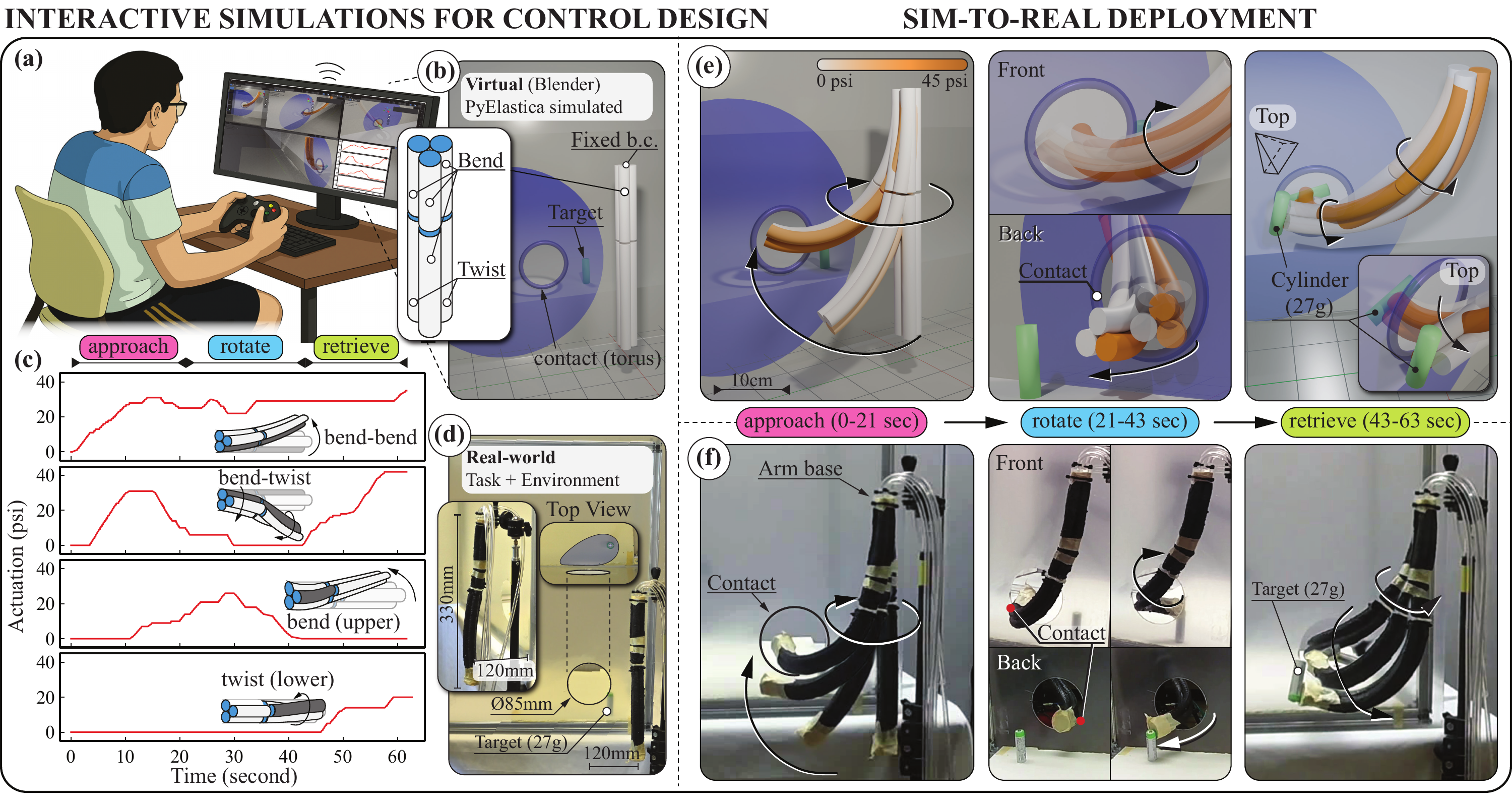}
\caption{
\textbf{Control strategies designed using interactive digital twins are subsequently transferred to physical hardware.}
(a) Control sequences are interactively (human in-the-loop) identified using a B3-BR2 digital twin immersed in a virtual environment.
(b) Digital twin in its operating environment, visualized in Blender.
(c) Pressure inputs that allow target retrieval are identified by user in simulations.
(d) The real-world experimental setup mirroring the virtual setup. The blue area in the top view indicates the region within which the target payload (AA battery) was randomly placed and from which it was consistently retrieved.
(e) Simulated deformation sequence executing the approach-rotate-retrieve task.
(f) Physical actuation sequence of the B3-BR2 retrieving the battery. Videos in SI.
}
\vspace{-8pt}
\label{fig:forward-control}
\end{figure*}

\section{Interactive design of control strategies via digital twins embedded in a virtual environment}\label{sec:interactive-design-and-forward-control}

With the numerical framework established, we proceed with demonstrating the interactive design of control strategies in simulations, by immersing our SCA digital twins in a virtual environment. Obtained controllers are then transferred to a corresponding physical prototype for experimental verification.
Particularly for SCAs, where targeted applications involve unstructured, non-deterministic environments, virtual or digital twin planning strategies hold considerable promise for remote operation.

We consider a SCA configured in a B3-BR2 (Fig.~\ref{fig:forward-control}b) fashion, loosely inspired by the combination of shoulder and elbow joints in the human arm. The arm is tasked with reaching and retrieving a metallic object (AA battery) located behind an acrylic wall. The wall presents a small circular opening through which the arm must insert to reach the battery, which is misaligned relative to the opening. A magnet is attached to the SCA's tip, to securely hold the battery upon contact and allow its retrieval. This is a challenging task, as it requires the SCA to approach the opening with a posture conducive to insert, reach around the battery, and then extract the payload (which is relatively heavy and geometrically awkward) while contending with wall boundaries and environmental effects.

To streamline the computational design of a suitable actuation strategy, we integrate a joypad within our simulation framework. Inputs from the joypad are relayed to the framework and transformed into pressure inputs to simulate and visualize the resulting SCA’s behavior in real time (Fig.~\ref{fig:forward-control}a).
This setup allows users to interactively test control actions, observe the SCA’s simulated response on screen, develop a physical intuition, and adjust their strategy (Fig.~\ref{fig:forward-control}c) before logging the final control inputs to be deployed experimentally.

The action space of the B3-BR2 is characterized by four pressure inputs, mimicking the human arm's movements: bend-bend configuration for arm-raising, bend-twist configuration for shoulder adduction and medial rotation, bend (upper) for opposite adduction, and twist (lower) for opposite medial rotation (Fig.~\ref{fig:forward-control}c).
A middle connector is specifically designed to pair pressure inputs from the top segment to the neighboring bottom segment for bend-bend and bend-twist co-actuation.
To separately control the bend (upper) actuator and twist (lower) actuator without an extra tether at the joint, a tube is inserted inside the upper actuator to link the regulator from the arm base to the lower actuator.
This pairing strategy is informed by trial and error simulations, to provide sufficient flexibility for completing the task while reducing the action space (thus simplifying control).
Such approach may be replaced in the future with a formal optimization methodology.
We note that action space planning/reduction is a common challenge in multi-actuator control and robotic design, where action spaces of dimension lower than the number of physical actuators may be desirable.
Our simulation framework then allows us to strategically pair and couple deformations, informing SCA design and fabrication.

Prior to logging the control data from our interactive simulation environment, users are allowed to conduct multiple practices to become familiar with the joypad controls. Pressure inputs are filtered to limit accelerations and maximum pressure changes, preventing structural instabilities and ensuring that the system operates in accordance with the equipment actuation rates.
Thanks to the reduced action space and intuitive control interface, a user typically needs about half an hour to become proficient and consistently succeed in the battery-retrieval task.
Once the user converges on a control strategy, the corresponding action sequence and pressure inputs are logged.
Logged controls (Fig.~\ref{fig:forward-control}c) are then directly used to drive the real SCA, without any human intervention.

Snapshots of the simulated SCA during a successful reach-and-retrieval maneuver are illustrated in Fig.~\ref{fig:forward-control}e.
During the `approach' phase (0–21~s), combined bend–bend and bend–twist actuation lifts the SCA tip (from the rest configuration of Fig.~\ref{fig:forward-control}b) towards the opening location.
During the subsequent `rotate' (21–43~s) and `retrieve' (43–63~s) phases, the SCA slips in (`rotate' phase) and out (`retrieve' phase) of the opening, intermittently entering in contact with the circular boundaries. This plays an important role. The arm indeed leans on the edges of the entrance, using them to rectify potential control inaccuracies. In order to capture this interaction, in simulations contact is implemented via an Hertzian model \cite{Hertz:1882}. As the arm rotates into the opening, it finally reaches to the battery. Upon contact, the simulation is paused (time: 43~s) and a payload equivalent to the battery's weight is added to the arm tip. The transient magnetic attraction between arm and battery is not considered in simulations, however the battery cylindrical geometry is taken into account to handle potential collisions with the walls during retrieval. Finally, the SCA pulls the battery out of the opening by engaging the combination of bend (upper) and twist (lower) actuations (see SI Video).

Next, we replay the simulated, logged pressure controls into the physical robot (Fig.~\ref{fig:forward-control}f).
The compliance of the SCA allows it to insert through the opening, lean against its edges, and accommodate posture discrepancies relative to simulations. As the tip of the SCA rotates into place, it approaches the battery with the magnet securely holding onto it.
Finally, as predicted in simulations, the arm slips out of the opening, with its compliance again helping to overcome contacts and control inaccuracies, to then complete the task (see SI Video). 
The same motion sequence was tested over 20 trials, slightly varying the battery position, and the SCA was able to successfully complete the task in all instances.

These results confirm the viability of our simulations, leading to predictable and reliable experimental execution.

\section{Conclusion}

In order to accelerate soft robotic deployment to real-world applications, design, prototyping, and control strategies tailored to their highly non-linear nature are necessary.
Towards a comprehensive rational design approach, fabrication modularity and reliable numerical simulations are key aspects.
Here, a Cosserat modeling approach is developed for the direct numerical simulation of soft manipulators realized through the modular composition of FREE actuators.
Our simulator reflects experimental modularity by allowing the user to connect rods into complex architectures.
By incorporating FREEs' specific physical parameters and actuation mechanisms into our rods representation, obtained digital twins are shown to recover experimental deformations of FREE-composed arms of increasing complexity. 
Concurrently, to facilitate the collection of such data, a dedicated pre-processing stage is integrated in the motion tracking pipeline of the Vicon system, addressing issues of reliability and accuracy associated with highly deformable, small-volume soft slender robots.

Finally, to illustrate the utility of our models for designing control strategies, we render our simulations interactive via the integration of a joypad system. Through the joypad, pressure inputs to the digital twins are online, manually controlled. A user is then challenged to drive a virtual robot arm through a wall opening, reach to a hidden object, and retrieve it. This task is specifically designed to test robust handling of contact and loading conditions in constrained environments, both in simulation and experiments. Joypad commands, and corresponding pressure inputs, are logged and subsequently played back to the physical system after which the simulation is modeled. Upon the automated execution of the interactively designed control strategy, the real arm is found to consistently and successfully retrieve the object over tens of runs.

Overall, this work advances the effective design, simulation, and interactive control of modular soft arms amidst uncertain environmental contacts and forces, paving the way to human-in-the-loop testing, tele-operation, iterative actuation programming, and motion validation under real-world uncertainties.

\bibliographystyle{IEEEtran}
\bibliography{ICRA2024-PyElastica}


\newpage
\section*{Appendix}\label{appendix}

Here we report key system parameters and simulation errors relative to experiments. Experimental data are available upon request. Simulation code is available at: 

\begin{table}[h!]
\centering
\begin{tabular}{|ll|}
\hline
\multicolumn{2}{|c|}{\textbf{Material Parameters}}                                                                                                                                                       \\ \hline
\multicolumn{1}{|l|}{\begin{tabular}[c]{@{}l@{}}density $\left(\frac{\text{kg}}{\text{m}^3}\right)$ (latex/rubber)\end{tabular}} & 1000                                                                \\ \hline
\multicolumn{1}{|l|}{length $\left(\text{cm}\right)$}                                                                              & 12 or 18                                                            \\ \hline
\multicolumn{1}{|l|}{radius $\left(\text{cm}\right)$}                                                                              & \begin{tabular}[c]{@{}l@{}}8.52 (outer)\\ 4.76 (inner)\end{tabular} \\ \hline
\multicolumn{1}{|l|}{Youngs modulus $\left(\text{MPa}\right)$}                                                                     & 1.5                                                                 \\ \hline
\multicolumn{1}{|l|}{Poisson ratio}                                                                                                & 0.5                                                                 \\ \hline
\multicolumn{1}{|l|}{actuation calibration factor $\gamma$}                                                                        & 0.85 - 1.15                                                         \\ \hline
\multicolumn{2}{|c|}{\textbf{FREE Parameters}}                                                                                                                                                           \\ \hline
\multicolumn{1}{|l|}{winding number}                                                                                               & \begin{tabular}[c]{@{}l@{}}1 (bending)\\ 2 (rotation)\end{tabular}  \\ \hline
\multicolumn{1}{|l|}{actuation range $\left(\text{psi}\right)$}                                                                    & 0 - 45                                                              \\ \hline
\end{tabular}
\caption{Key parameters} \label{table:param}
\end{table}

\begin{table}[h!]
\centering
\begin{tabular}{|ccccc|}
\hline
\multicolumn{1}{|c|}{\textbf{}} &
  \multicolumn{1}{c|}{\textbf{Tip~Position}} &
  \multicolumn{1}{c|}{\textbf{\begin{tabular}[c]{@{}c@{}}Total\\ Twist\end{tabular}}} &
  \multicolumn{1}{c|}{\textbf{\begin{tabular}[c]{@{}c@{}}Total\\ Bending\end{tabular}}} &
  \textbf{Elongation} \\ \hline
\multicolumn{5}{|c|}{\textbf{Twist-Twist}}                                                                                           \\ \hline
\multicolumn{1}{|c|}{\textbf{10 psi}} & \multicolumn{1}{c|}{2.1\%} & \multicolumn{1}{c|}{3.0\%} & \multicolumn{1}{c|}{N/A}   & 0.7\% \\ \hline
\multicolumn{1}{|c|}{\textbf{20 psi}} & \multicolumn{1}{c|}{5.9\%} & \multicolumn{1}{c|}{4.6\%} & \multicolumn{1}{c|}{N/A}   & 2.3\% \\ \hline
\multicolumn{5}{|c|}{\textbf{Bend-Bend}}                                                                                             \\ \hline
\multicolumn{1}{|c|}{\textbf{10 psi}} & \multicolumn{1}{c|}{3.5\%} & \multicolumn{1}{c|}{N/A}   & \multicolumn{1}{c|}{0.4\%} & 2.5\% \\ \hline
\multicolumn{1}{|c|}{\textbf{25 psi}} & \multicolumn{1}{c|}{4.3\%} & \multicolumn{1}{c|}{N/A}   & \multicolumn{1}{c|}{6.2\%} & 6.9\% \\ \hline
\multicolumn{5}{|c|}{\textbf{BR2-B3}}                                                                                                \\ \hline
\multicolumn{1}{|c|}{\textbf{pose 1}} & \multicolumn{1}{c|}{4.5\%} & \multicolumn{1}{c|}{3.1\%} & \multicolumn{1}{c|}{8.2\%} & 3.1\% \\ \hline
\multicolumn{1}{|c|}{\textbf{pose 2}} & \multicolumn{1}{c|}{7.2\%} & \multicolumn{1}{c|}{4.5\%} & \multicolumn{1}{c|}{5.2\%} & 4.8\% \\ \hline
\multicolumn{1}{|c|}{\textbf{pose 3}} & \multicolumn{1}{c|}{8.2\%} & \multicolumn{1}{c|}{3.7\%} & \multicolumn{1}{c|}{5.6\%} & 2.4\% \\ \hline
\end{tabular}
\caption{Comparison simulations vs. experiments using relative percent error $\left[e_\text{relative}(A, B)=\tfrac{|A-B|}{|B|}\times100\right]$. We evaluated integrated quantities that characterize the deformation of the entire rod: tip position: $\position(L)$, total twist~\cite{Charles2019Topology}: $\left[\int_0^L |\curvature_3(s)|\,ds\right]$, total bend \cite{Antman1995Nonelinear}: $\left[\int_0^L \sqrt{\curvature_1^2+\curvature_2^2}\,ds\right]$, and total elongation: $\left[\int_0^L |\shear_3-1|\,ds\right]$.}
\label{tab:accuracy}
\end{table}




\end{document}